\let\originalbibliographystyle\bibliographystyle
\renewcommand{\bibliographystyle}[1]{\originalbibliographystyle{bst/#1}}
\documentclass[pdflatex,sn-nature]{sn-jnl}% Nature reference style
\usepackage{graphicx}%
\usepackage{multirow}%
\usepackage{amsmath,amssymb,amsfonts}%
\usepackage{amsthm}%
\usepackage{mathrsfs}%
\usepackage[title]{appendix}%
\usepackage{xcolor}%
\usepackage{textcomp}%
\usepackage{manyfoot}%
\usepackage{booktabs}%
\usepackage{algorithm}%
\usepackage{algorithmicx}%
\usepackage{algpseudocode}%
\usepackage{listings}%
\theoremstyle{thmstyleone}%
\newtheorem{proposition}{Proposition}% 

\theoremstyle{thmstyletwo}%
\theoremstyle{thmstylethree}%
\usepackage{array}
\usepackage{stmaryrd}
\usepackage{bm}
\usepackage{gensymb}
\usepackage{siunitx}
\begin{document}

\title{Initialization-Free LEO Doppler Positioning via Semidefinite Relaxation: A Cold-Start Approach}

% Author order, affiliations and contact details supplied by the authors.
\author*[1]{\fnm{Baoshan} \sur{Song}}
\email{baoshan.song@connect.polyu.hk}

\author[1]{\fnm{Weisong} \sur{Wen}}
\email{welson.wen@polyu.edu.hk}

\author[1]{\fnm{Qi} \sur{Zhang}}
\email{qi915.zhang@connect.polyu.hk}

\author[1]{\fnm{Bing} \sur{Xu}}
\email{pbing.xu@polyu.edu.hk}

\author[1]{\fnm{Penggao} \sur{Yan}}
\email{penggao.yan@polyu.edu.hk}

\author[1]{\fnm{Yuan} \sur{Li}}
\email{23036326R@connect.polyu.hk}

\author*[1]{\fnm{Li-Ta} \sur{Hsu}}
\email{lt.hsu@polyu.edu.hk}

\affil[1]{%
  \orgdiv{Department of Aeronautical and Aviation Engineering},
  \orgname{The Hong Kong Polytechnic University},
  \orgaddress{\city{Hong Kong}, \country{China}}
}

\abstract{Low Earth Orbit (LEO) satellite Doppler measurements offer an alternative source of positioning information for Internet of Things devices when Global Navigation Satellite System signals are unavailable. Conventional local optimisation methods depend on an accurate initial position and can fail during a complete cold start. We propose an initialization-free LEO Doppler positioning framework based on semidefinite relaxation. The method reformulates the nonlinear positioning problem into a semidefinite programme to obtain a receiver position estimate without prior location knowledge. We analyse relaxation tightness under noiseless conditions and derive sufficient noise bounds under parameter disturbances. Simulations and field experiments using opportunistic Iridium-NEXT signals evaluate initialization sensitivity, positioning accuracy and computational cost. In the field experiment, the proposed method achieves a three-dimensional positioning error of 140 m without an initial position, whereas local solvers converge to poor solutions under large initial offsets. Using the semidefinite estimate to initialise Gauss-Newton or Dog-Leg reduces the positioning error to 130 m. The framework provides a cold-start initialization stage for subsequent local refinement in opportunistic LEO positioning. The implementation and datasets are publicly available to support reproducibility and further research.}


\maketitle

\section{Introduction}\label{sec:introduction}

\textbf{\textit{LEO satellites serve as a vital alternative and supplement to GNSS for IoT PNT services.}} 
Position, Navigation, and Timing (PNT) services\cite{tan_toward_2025} are foundational to modern technological infrastructures and Internet of Things (IoT) ecosystems, enabling location-based services, autonomous mobility, and distributed sensor networks\cite{liu_opposiiton_2026}. 
For decades, Global Navigation Satellite Systems (GNSS) have been the primary provider of PNT. 
However, conventional GNSS signals face severe vulnerabilities in IoT deployments, including urban signal blockage\cite{wen_3d_2023}, intentional jamming\cite{wang_gnss_2018}, and spoofing attacks\cite{bai_gnss_2024}. 
GNSS disruptions can cause widespread operational failures and multi-billion-dollar economic losses\cite{kassas_resilient_2022}. 
To mitigate these risks, Low Earth Orbit (LEO) satellites have emerged as a promising alternative and augmentation source for GNSS-denied PNT, benefiting from higher received signal strength, rapid geometric diversity, and dense constellation coverage\cite{khalife_carrier_2021}. 

By the end of 2023, there were over 8,400 active LEO satellites in space, offering vast potential for resilient IoT positioning. 
These satellites generally fall into two categories: 
(1) \textit{Dedicated navigation satellites} equipped with specialized payloads and customized PNT signals (e.g., Xona Space Systems' PULSAR constellation\cite{miller_snap_2023}), though their total number remains currently limited; and 
(2) \textit{Non-navigation satellites} primarily deployed for broadband communication and remote sensing, such as Starlink, OneWeb, and Orbcomm. 
Because non-navigation satellites do not transmit open navigation messages (such as standard pseudoranges or broadcast ephemerides) to unauthorized users\cite{deng_noncooperative_2022}, unassisted IoT nodes cannot perform standard GNSS-like trilateration. 
Fortunately, opportunistic navigation framework allows IoT receivers to leverage LEO signals of opportunity (SOPs) for positioning\cite{kassas_resilient_2022, khalife_first_2021}. 
Among various SOP observables, the carrier Doppler shift stands out as a primary, readily accessible product derived during signal acquisition and tracking\cite{wang_doppler_2023}. 
Considering the massive potential of LEO mega-constellations, this paper focuses on developing an opportunistic LEO Doppler positioning framework for IoT applications.

\textbf{\textit{LEO Doppler positioning is effective but requires a reliable initial estimate.}} 
Among various LEO SOP measurements, carrier Doppler shift is widely used because it can be directly obtained during signal tracking\cite{guo_instantaneous_2023}. 
Although Doppler navigation dates back to the Transit system\cite{stansell_transit_1971}, the rapid growth of mega-LEO constellations has brought new interest to this field\cite{mcdowell_low_2020}. 
To use non-cooperative LEO SOPs, several unknown parameters such as signal structures, clock biases, and ephemerides must be handled, and existing works have proposed effective methods to solve these issues\cite{kassas_resilient_2022, humphreys_signal_2023}. 
For instance, recent research introduced a warm start framework using a marginalized particle filter to estimate receiver position and LEO orbital errors simultaneously when a rough initial guess is available\cite{kassas_warmstart_2024}. 
When a coarse position guess is available, these local refinement algorithms and filters work fast and accurately. 
However, due to the non-convex nature of Doppler positioning, local methods easily get trapped in local minima if the initial guess is too far off. 
For example, previous study\cite{shi_revisiting_2023} shows that for LEO satellites at 550~km altitude, the initial error should be within 300~km to ensure convergence. 
In a complete cold-start scenario where no prior position is given, obtaining such an initial guess is difficult. 
To support existing local optimization solvers and filter-based warm start frameworks, an initialization-free cold-start method is needed to provide a reliable initial position without any prior knowledge.

\textbf{\textit{Convex optimization offers a promising initialization-free approach.}} 
With recent advances in optimization theory, convex relaxation methods can solve certain non-convex problems reliably without requiring an initial estimate\cite{boyd_convex_2004}. 
By applying Lagrangian duality, an original non-convex problem can be relaxed into a convex semidefinite programming problem. 
Solving this convex dual problem yields a lower bound for the original objective function value. 
The difference between the original and relaxed objective values is known as the duality gap. 
Under specific conditions, when the duality gap is zero, the exact solution of the original non-convex problem can be recovered. 
Such convex relaxation techniques have been successfully applied in various robotics and positioning tasks, including geometric registration\cite{yang_teaser_2020}, sensor calibration\cite{giamou_certifiably_2019,song_certifiable_2026}, simultaneous localization and mapping\cite{rosen_se-sync_2019}, rotation synchronization\cite{liu_resync_2023}, camera placement\cite{kaveti_oasis_2024}, and wireless localization\cite{yan_robust_2022}. 
However, applying semidefinite relaxation to the LEO Doppler positioning problem to provide a reliable cold-start initialization remains under-explored.

\begin{figure}[htbp]
\centering
\begin{minipage}{0.96\linewidth}
\centering
\textbf{(a)}\\[2pt]
\includegraphics[width=\linewidth,height=0.31\textheight,keepaspectratio]{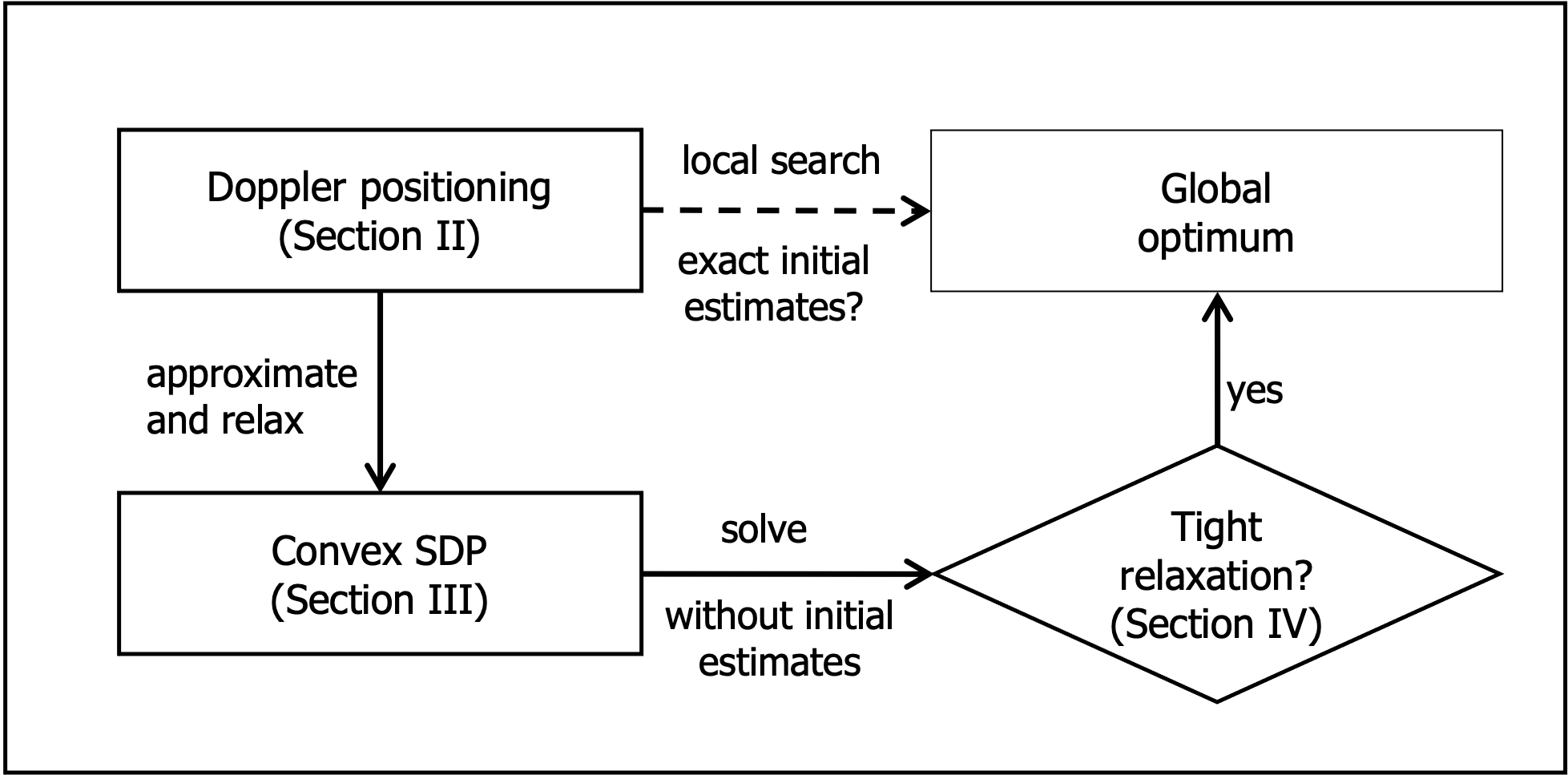}
\end{minipage}
\par\medskip
\begin{minipage}{0.65\linewidth}
\centering
\textbf{(b)}\\[2pt]
\includegraphics[width=\linewidth,height=0.21\textheight,keepaspectratio]{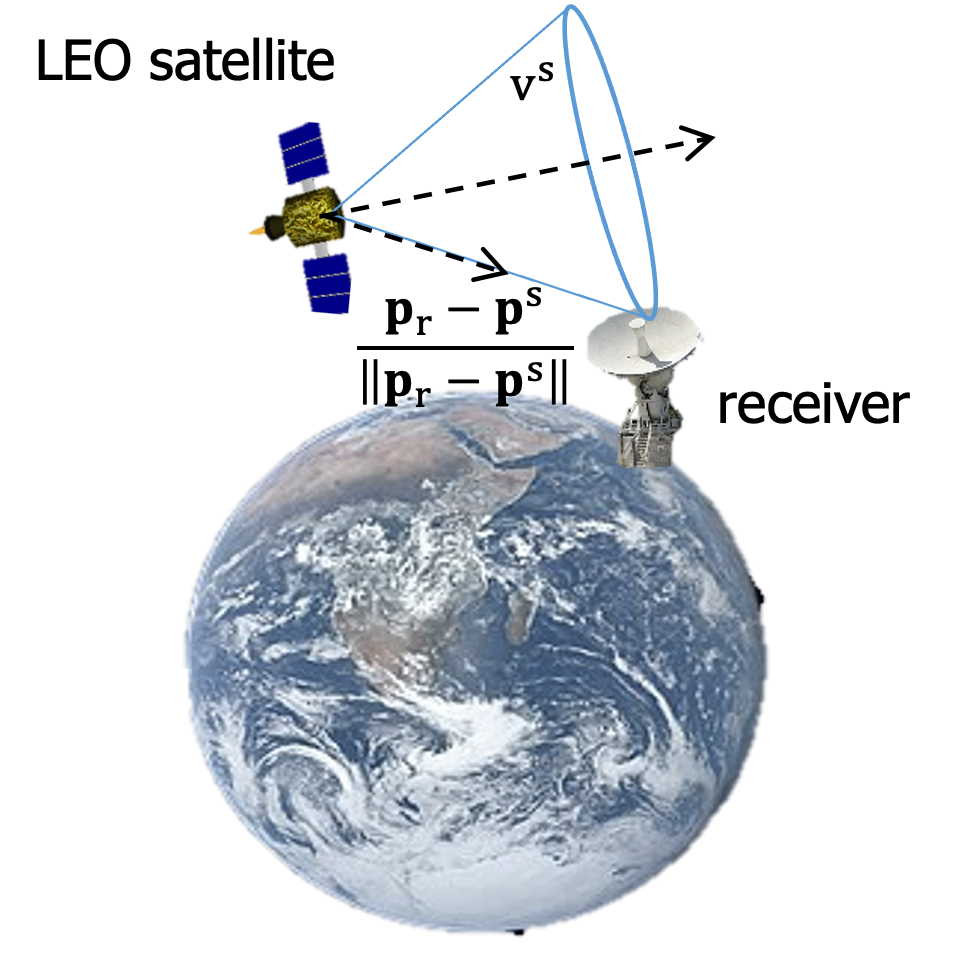}
\end{minipage}
\par
\caption{Framework and measurement geometry. (a) Overview of the proposed initialization-free LEO Doppler positioning framework. Section \ref{sec:problem} presents the Doppler positioning problem. Section \ref{sec:sdp_method} formulates the SDP relaxation that solves the problem without initial estimates. Section \ref{sec:guarantee} analyzes the tight relaxation conditions to guarantee exact recovery. (b) Illustration of LEO Doppler positioning geometry. The carrier Doppler shift is determined by the projection of the satellite velocity vector \(\mathbf{v}^s\) onto the line-of-sight unit vector \(\frac{\mathbf{p}_r - \mathbf{p}^s}{\|\mathbf{p}_r - \mathbf{p}^s\|}\) between the receiver position \(\mathbf{p}_r\) and the satellite position \(\mathbf{p}^s\).}
\label{fig:framework}
\label{fig:doppler_positioning}
\end{figure}

To address this gap, this paper presents an initialization-free LEO Doppler positioning framework based on convex optimization. 
This work builds upon our previous certifiable Doppler-based orbit determination method\cite{song_certifiably_2025}, extending the convex optimization formulation to ground receiver positioning using opportunistic LEO Doppler measurements. 
While our prior work focused on satellite orbit estimation from ground stations, this paper addresses the inverse problem of determining ground receiver coordinates from non-cooperative LEO satellites without any initial position guess.

The main contributions of this paper are summarized as follows.

\begin{itemize}
    \item \textbf{Convex relaxation framework for LEO Doppler cold start.} 
    We formulate the non-convex LEO Doppler positioning problem into a semidefinite programming relaxation. This approach allows us to obtain a reliable initial receiver position estimate using standard convex solvers without requiring prior spatial knowledge.

    \item \textbf{Theoretical analysis of relaxation tightness and noise sensitivity.} 
    We analyze the conditions under which the semidefinite relaxation is tight in ideal noiseless scenarios. In addition, we examine the noise bound behavior under parameter disturbances to evaluate the robustness of the relaxed formulation.

    \item \textbf{Comprehensive experimental evaluation and open-source contribution.} 
    We evaluate the proposed cold-start method through simulations and real-world field experiments using Iridium-NEXT LEO satellite signals. Experimental results demonstrate that the SDP-based estimate serves as an effective initializer for local search methods under large initial errors. To support reproducibility, all code and datasets are open sourced (see the Data availability and Code availability statements).
\end{itemize}

The remainder of this paper is organized as follows. 
Section~\ref{sec:problem} introduces the mathematical model for LEO Doppler positioning and analyzes the cold-start challenge. 
Section~\ref{sec:sdp_method} derives the proposed SDP-based cold-start positioning method using semidefinite relaxation. 
Section~\ref{sec:guarantee} presents the theoretical performance analysis, including relaxation tightness and noise sensitivity bounds for the relaxed formulation. 
Section~\ref{sec:experiments} evaluates the performance of the proposed framework through numerical simulations and real-world field experiments. 
Finally, Section~\ref{sec:conclusion} concludes this paper and outlines future research directions.

\textit{Notation and Abbreviations}: 
Lowercase and uppercase boldface letters denote vectors and matrices, respectively, such as $\mathbf{a}$ and $\mathbf{A}$. 
$\text{diag}\{a_1, \dots, a_n\}$ represents a diagonal matrix with diagonal elements $a_1, \dots, a_n$. 
$\mathbf{a}_{(i:j)}$ denotes a vector containing the $i$-th to $j$-th components of $\mathbf{a}$, while $\mathbf{A}_{(i:j)}$ denotes the submatrix formed by rows and columns from index $i$ to $j$. 
$\mathbf{I}_n$ denotes the $n \times n$ identity matrix. 
$\text{rank}(\mathbf{A})$, $\text{corank}(\mathbf{A})$, and $\text{tr}(\mathbf{A})$ represent the rank, corank, and trace of a square matrix $\mathbf{A}$, respectively. 
$\mathbb{R}^n$ and $\mathbb{S}^n$ denote the $n$-dimensional real vector space and the space of $n \times n$ symmetric matrices, respectively. 
$\mathbf{A} \succeq 0$ indicates that matrix $\mathbf{A}$ is symmetric positive semidefinite. 
The notation $\|\mathbf{a}\|$ and $\|\mathbf{A}\|$ denote the Euclidean norm of a vector $\mathbf{a}$ and the spectral norm of a matrix $\mathbf{A}$. 
The Frobenius inner product of two matrices $\mathbf{A}$ and $\mathbf{B}$ is denoted by $\mathbf{A} \bullet \mathbf{B} = \text{tr}(\mathbf{A}^T \mathbf{B})$. 

For clarity, key acronyms used in this paper are defined as follows: 
Position, Navigation, and Timing (PNT); 
Low Earth Orbit (LEO); 
Signals of Opportunity (SOP); 
Semidefinite Programming (SDP); 
Gauss-Newton (GN); 
Dog-Leg (DL); 
Software-Defined Radio (SDR); 
and Non-linear Weighted Least-Squares (NWLS).

\section{LEO Doppler Positioning Problem Formulation}\label{sec:problem}

Unless explicitly stated otherwise, all position and velocity vectors in this paper are defined in the Earth-Centered, Earth-Fixed (ECEF) reference frame. Consider a static ground receiver located at position $\mathbf{p}_r \in \mathbb{R}^3$ with zero velocity $\mathbf{v}_r = \mathbf{0}$. For $N$ opportunistic Doppler observations collected from overhead LEO satellites across single or multiple epochs, let $\mathbf{p}^s_i \in \mathbb{R}^3$ and $\mathbf{v}^s_i \in \mathbb{R}^3$ denote the position and velocity of the satellite for the $i$-th measurement, where $i \in \{1, \dots, N\}$.
After correcting atmospheric delays using standard environmental models\cite{shi_revisiting_2023}, the Doppler observation expressed in pseudorange rate is modeled as
\begin{equation}
    y_i = \frac{(\mathbf{p}_r - \mathbf{p}^s_i)^T (\mathbf{v}_r - \mathbf{v}^s_i)}{\|\mathbf{p}_r - \mathbf{p}^s_i\|} + c \cdot d\dot{t}_r + \varepsilon_i
    \label{equ:doppler_true}
\end{equation}
where $y_i$ is the measured pseudorange rate ($m/s$), $c$ is the speed of light ($m/s$), $d\dot{t}_r$ is the dimensionless receiver clock drift rate ($s/s$), and $\varepsilon_i \sim \mathcal{N}(0, \sigma_i^2)$ represents white Gaussian noise ($m/s$). Geometrically, each observation restricts the receiver position to a conical surface with the satellite at the vertex\cite{guo_instantaneous_2023}. Intersecting multiple conical surfaces yields the estimated receiver position.

Given $N$ pseudorange rate observations $\mathbf{y} = [y_1, \dots, y_N]^T \in \mathbb{R}^N$, static LEO Doppler positioning is formulated as a nonlinear weighted least-squares problem
\begin{equation}
    \min_{\mathbf{p}_r, d\dot{t}_r} \sum_{i=1}^N w_i \left( y_i - \frac{(\mathbf{p}_r - \mathbf{p}^s_i)^T \mathbf{v}^s_i}{\|\mathbf{p}_r - \mathbf{p}^s_i\|} - c \cdot d\dot{t}_r \right)^2
    \label{equ:wls}
\end{equation}
where $w_i > 0$ denotes the measurement weighting factor inversely proportional to the noise variance $\sigma_i^2$.

In numerical optimization, a function or feasible set is non-convex if line segments connecting pairs of valid points do not lie entirely within the function graph or set boundary, creating multiple local minima\cite{boyd_convex_2004}. Unlike convex problems that admit guaranteed global convergence, problem (\ref{equ:wls}) is non-convex primarily due to the non-linear fractional range vector term $(\mathbf{p}_r - \mathbf{p}^s_i) / \|\mathbf{p}_r - \mathbf{p}^s_i\|$ and the bilateral product of unknown position coordinates. Traditional local search methods such as Gauss-Newton approximate (\ref{equ:wls}) via local linearization, but they inherently depend on precise initial estimates and cannot guarantee convergence when initialized far from the true position\cite{gratton_approximate_2007}. To overcome this cold-start limitation, the next section reformulates (\ref{equ:wls}) using convex relaxation to eliminate initial value dependence.

\section{SDP-Based Cold-Start Positioning Method} \label{sec:sdp_method}

While local search methods are straightforward to set up, they often fail without an accurate initial position. Conversely, convex relaxation methods reformulate the original problem into a convex optimization framework, enabling reliable global resolution without requiring initial guesses\cite{boyd_convex_2004}. 
To overcome the cold-start limitation, this section develops an initialization-free positioning framework using SDP relaxation. The derivation follows three key steps. First, we remove the fractional term in (\ref{equ:wls}) by multiplying the range residual with the geometric distance, transforming the non-convex fractional model into a polynomial optimization problem (POP). Second, we introduce auxiliary variables to lift the POP into a quadratically constrained quadratic program (QCQP). Third, we drop the non-convex rank-one constraint to obtain a solvable semidefinite relaxation.

\subsection{Polynomial Optimization Formulation}\label{sec:pop}

The primary source of non-convexity in the original NWLS formulation (\ref{equ:wls}) stems from the fractional line-of-sight range vector $(\mathbf{p}_r - \mathbf{p}^s_i) / \|\mathbf{p}_r - \mathbf{p}^s_i\|$. To eliminate this fractional dependency, we multiply the $i$-th residual in (\ref{equ:wls}) by the geometric distance $\rho_i = \|\mathbf{p}_r - \mathbf{p}^s_i\|$, transforming the formulation into
\begin{equation}
    \min_{\mathbf{p}_r, d\dot{t}_r} \sum_{i=1}^N \tilde{w}_i \left( \rho_i (y_i - c \cdot d\dot{t}_r) - (\mathbf{p}_r - \mathbf{p}^s_i)^T \mathbf{v}^s_i \right)^2
    \label{equ:pop}
\end{equation}
where $\tilde{w}_i$ denotes the modified weighting factor. 
Because multiplying the residual by $\rho_i$ scales the measurement noise variance by $\rho_i^2$, statistical consistency with the original NWLS formulation suggests $\tilde{w}_i = w_i / \rho_i^2 = 1 / (\sigma_i^2 \rho_i^2)$. Since the primary objective of this formulation is to deliver a reliable cold-start estimate rather than high-precision refinement, an approximate geometric distance $\rho_i$ is sufficient to compute $\tilde{w}_i$. Following established range-dependent least-squares literature\cite{chan_simple_1994}, an iterative weighting update can be applied using the updated position estimate. Because orbital distances are large and vary smoothly, a coarse range approximation rapidly yields a high-quality initial position estimate within iterations. This estimate successfully pulls the solution out of local minima traps, leaving fine positioning refinement to subsequent local optimization solvers.

\subsection{QCQP Formulation}\label{sec:qcqp}

Although polynomial optimization problems can be addressed directly, solving general polynomial programs is computationally expensive\cite{yang_polynomial-time_2019}. To enable tractable polynomial-time convex relaxation, we reformulate the polynomial objective into a quadratically constrained quadratic program (QCQP). 
Following the variable lifting paradigm\cite{dumbgen_globally_2024}, we introduce auxiliary higher-dimensional variables to linearize the higher-degree monomial terms in (\ref{equ:pop}). The lifted state vector $\mathbf{y} \in \mathbb{R}^{2N+4}$ is defined as
\begin{equation}
    \mathbf{y} = [\mathbf{p}_r^T, c \cdot d\dot{t}_r, \rho_1, \dots, \rho_N, z_1, \dots, z_N]^T
    \label{equ:lifted_state}
\end{equation}
where $z_i = c \cdot \rho_i \cdot d\dot{t}_r$ for $i \in \{1, \dots, N\}$ linearizes the bilinear product between geometric distance and receiver clock drift rate. To preserve the algebraic consistency among the lifted variables, we enforce quadratic redundant constraints corresponding to the geometric range $\rho_i = \|\mathbf{p}_r - \mathbf{p}_i^s\|$ and the auxiliary product $z_i$. 

Let $\mathbf{Y} = \mathbf{y}\mathbf{y}^T \in \mathbb{S}^{2N+4}_+$ denote the outer product matrix of the lifted state vector. The original optimization problem (\ref{equ:pop}) is equivalently reformulated into the following QCQP
\begin{equation}
\begin{array}{ll}
    \min_{\mathbf{y}, \mathbf{Y}} & \text{tr}(\mathbf{A}^T \mathbf{W} \mathbf{A} \mathbf{Y}) + 2 \mathbf{k}^T \mathbf{W} \mathbf{A} \mathbf{y} + \mathbf{k}^T \mathbf{W} \mathbf{k} \\
    \text{s.t.} & \mathbf{y}_{(4+i)}^2 = \text{tr}(\mathbf{Y}_{(1:3)}) - 2 (\mathbf{p}_i^s)^T \mathbf{y}_{(1:3)} + (\mathbf{p}_i^s)^T \mathbf{p}_i^s \\
                & \mathbf{y}_{(4+N+i)} = \mathbf{Y}_{(4, 4+i)} \\
                & \mathbf{Y} = \mathbf{y}\mathbf{y}^T, \quad i \in \{1, \dots, N\}
\end{array} 
\label{equ:qcqp}
\end{equation}
where $\mathbf{W} = \text{diag}\{\tilde{w}_1, \dots, \tilde{w}_N\} \in \mathbb{S}^N_+$ contains the modified residual weights. The coefficient matrix $\mathbf{A} \in \mathbb{R}^{N \times (2N+4)}$ and vector $\mathbf{k} \in \mathbb{R}^N$ are defined as
\begin{equation}
    \mathbf{A} = \begin{bmatrix}
        (\mathbf{v}_1^s)^T & 0 & y_1 & \dots & 0 & -1 & \dots & 0 \\
        \vdots & \vdots & \vdots & \ddots & \vdots & \vdots & \ddots & \vdots \\
        (\mathbf{v}_N^s)^T & 0 & 0 & \dots & y_N & 0 & \dots & -1
    \end{bmatrix}
\end{equation}
\begin{equation}
    \mathbf{k} = [ -(\mathbf{p}_1^s)^T \mathbf{v}_1^s, \dots, -(\mathbf{p}_N^s)^T \mathbf{v}_N^s ]^T
\end{equation}

In formulation (\ref{equ:qcqp}), all non-convexity is now concentrated within the quadratic equality constraint $\mathbf{Y} = \mathbf{y}\mathbf{y}^T$. This equivalent matrix rank condition sets up the final step for semidefinite relaxation.

\subsection{Semidefinite Relaxation}\label{sec:sdp}

Because the equality constraint $\mathbf{Y} = \mathbf{y}\mathbf{y}^T$ in (\ref{equ:qcqp}) remains non-convex, we apply semidefinite relaxation to obtain a tractable convex formulation. By introducing the augmented matrix $\mathbf{S}$, the outer product condition $\mathbf{Y} = \mathbf{y}\mathbf{y}^T$ is equivalently represented as
\begin{equation}
    \mathbf{S} = \begin{bmatrix}
        \mathbf{Y} & \mathbf{y} \\
        \mathbf{y}^T & 1
    \end{bmatrix} \succeq 0, \quad \text{rank}(\mathbf{S}) = 1
    \label{con:rank_relax}
\end{equation}

The semidefinite relaxation scheme drops the non-convex rank-one constraint $\text{rank}(\mathbf{S}) = 1$, yielding the following semidefinite program (SDP) primal problem
\begin{equation}
\begin{array}{ll}
    \min_{\mathbf{S} \in \mathbb{S}^{2N+5}} & \mathbf{F} \bullet \mathbf{Y} + \mathbf{l}_0^T \mathbf{y} + c_0 \\
    \text{s.t.} & \mathbf{S} = \begin{bmatrix} \mathbf{Y} & \mathbf{y} \\ \mathbf{y}^T & 1 \end{bmatrix} \succeq 0 \\
                & \mathbf{G}_i \bullet \mathbf{Y} + \mathbf{l}_i^T \mathbf{y} + c_i = 0, \quad i \in \{1, \dots, 2N\}
\end{array} 
\label{equ:sdp_primal}
\end{equation}
where $\mathbf{F} = \mathbf{A}^T \mathbf{W} \mathbf{A}$, $\mathbf{l}_0 = 2 \mathbf{A}^T \mathbf{W} \mathbf{k}$, $c_0 = \mathbf{k}^T \mathbf{W} \mathbf{k}$, and $\mathbf{W} = \text{diag}\{\tilde{w}_1, \dots, \tilde{w}_N\}$. The constraint matrices and vectors for $i \in \{1, \dots, N\}$ are structured as
\begin{align}
    \mathbf{G}_i &= \text{diag}\{ \mathbf{1}_3, 0, -\mathbf{e}_i, \mathbf{0}_N \}, \label{equ:gi} \\
    \mathbf{l}_i &= [ -2(\mathbf{p}_i^s)^T, \mathbf{0}_{1 \times (2N+1)} ]^T, \quad c_i = (\mathbf{p}_i^s)^T \mathbf{p}_i^s, \notag \\
    \mathbf{G}_{i+N} &= \text{diag}\{ \mathbf{0}_{3\times 3}, \mathbf{E}_i, \mathbf{0}_{N \times N} \}, \label{equ:gi_n} \\
    \mathbf{l}_{i+N} &= [ \mathbf{0}_{1 \times 4}, \mathbf{e}_i^T, 0 ]^T, \quad c_{i+N} = 0 \notag
\end{align}
Here $\mathbf{E}_i \in \mathbb{S}^{N+1}$ contains non-zero block $\frac{1}{2} (\mathbf{e}_i \mathbf{e}_1^T + \mathbf{e}_1 \mathbf{e}_i^T)$, $\mathbf{1}_3$ is a $3$-dimensional all-ones vector, $\mathbf{0}_N$ is an $N$-dimensional zero vector, and $\mathbf{e}_i$ denotes the $i$-th standard basis vector in $\mathbb{R}^N$.

By dropping the rank constraint, formulation (\ref{equ:sdp_primal}) becomes a standard convex SDP that can be efficiently solved using off-the-shelf interior-point solvers (e.g., MOSEK). The solution to (\ref{equ:sdp_primal}) provides a reliable lower-bound and yields a high-quality cold-start estimate for ground receiver positioning.

\section{Theoretical Performance Guarantees}\label{sec:guarantee}

The semidefinite relaxation derived in Section \ref{sec:sdp} provides a convex lower-bound for the original non-convex LEO Doppler positioning formulation. To evaluate the quality of this convex relaxation, it is essential to analyze the duality gap and establish conditions under which the SDP relaxation remains tight. In this section, we apply Lagrangian duality to formulate the dual problem and examine the tightness behavior under both noiseless and noisy observation scenarios.

\subsection{Lagrangian Duality Formulation}\label{sec:duality}

Following the homogeneous dual matrix formulation\cite{dumbgen_globally_2024}, we construct the dual problem of the primal SDP (\ref{equ:sdp_primal}). Let $\boldsymbol{\lambda} = [\lambda_1, \dots, \lambda_{2N}]^T \in \mathbb{R}^{2N}$ denote the vector of Lagrangian multipliers associated with the equality constraints in (\ref{equ:sdp_primal}). The corresponding dual problem is expressed as
\begin{equation}
\begin{array}{ll}
    \max_{\boldsymbol{\lambda} \in \mathbb{R}^{2N}} & \gamma \\
    \text{s.t.} & \mathbf{H}(\boldsymbol{\lambda}, \gamma) \succeq 0
\end{array} 
\label{equ:sdp_dual}
\end{equation}
where the dual constraint matrix $\mathbf{H}(\boldsymbol{\lambda}, \gamma) \in \mathbb{S}^{2N+5}_+$ is linear in the multipliers $\boldsymbol{\lambda}$ and scalar dual variable $\gamma$, defined as
\begin{equation}
    \mathbf{H}(\boldsymbol{\lambda}, \gamma) = \begin{bmatrix}
        c_0 - \gamma & \frac{1}{2} \mathbf{l}_0^T \\
        \frac{1}{2} \mathbf{l}_0 & \mathbf{F}
    \end{bmatrix} + \sum_{i=1}^{2N} \lambda_i \begin{bmatrix}
        c_i & \frac{1}{2} \mathbf{l}_i^T \\
        \frac{1}{2} \mathbf{l}_i & \mathbf{G}_i
    \end{bmatrix}
    \label{equ:dual_matrix}
\end{equation}

By weak duality, any dual feasible point $(\boldsymbol{\lambda}, \gamma)$ satisfying $\mathbf{H}(\boldsymbol{\lambda}, \gamma) \succeq 0$ establishes a certified lower bound $\gamma$ for the primal objective cost in (\ref{equ:sdp_primal}). When strong duality holds and the duality gap vanishes, the rank-one condition of the primal matrix $\mathbf{S}$ can be verified through the nullspace structure of the dual matrix $\mathbf{H}(\boldsymbol{\lambda}, \gamma)$.

\subsection{Tightness in a Noiseless Scenario} \label{sec:noiseless}

Let $q^*$, $p^*$, and $d^*$ denote the optimal objective costs of the original QCQP (\ref{equ:qcqp}), primal SDP (\ref{equ:sdp_primal}), and dual SDP (\ref{equ:sdp_dual}), respectively, satisfying weak duality $q^* \ge p^* \ge d^*$. When zero duality gap holds ($p^* = d^*$) and the primal optimal solution satisfies $\text{rank}(\mathbf{Y}) = 1$, the relaxation is tight, allowing exact recovery of the positioning solution.
To analyze the tightness condition, we apply the sufficient optimality conditions from \cite{cifuentes_local_2022}. The sufficient conditions for zero duality gap between primal SDP (\ref{equ:sdp_primal}) and dual SDP (\ref{equ:sdp_dual}) are expressed as
\begin{equation}
    g_i(\mathbf{y}) = 0, \quad \mathbf{H}(\boldsymbol{\lambda}) \succeq 0, \quad \mathbf{H}(\boldsymbol{\lambda}) [1, \mathbf{y}^T]^T = \mathbf{0}
    \label{equ:kkt_conditions}
\end{equation}
where $g_i(\mathbf{y}) = 0$ represents primal feasibility for $i \in \{1, \dots, 2N\}$. By complementary slackness, the rank-one condition $\text{rank}(\mathbf{Y}) = 1$ is guaranteed when the dual matrix $\mathbf{H}(\boldsymbol{\lambda})$ has a corank of one\cite{dumbgen_globally_2024}. Specifically, because $\text{rank}(\mathbf{F}) = \text{rank}(\mathbf{A}) = N+3$ and the column dimension of $\mathbf{F}$ is $N+4$, we obtain $\text{corank}(\mathbf{F}) = (N+4) - (N+3) = 1$. Similarly, $\text{corank}(\mathbf{G}_i) = 1$ holds for constraint matrices. Consequently, $\mathbf{H}(\boldsymbol{\lambda})$ possesses a corank of one, certifying that the principal eigenvector of $\mathbf{S}$ yields the unique global minimizer of (\ref{equ:qcqp}) in the absence of noise.

\subsection{Tightness and Robustness under Noisy Measurements}\label{sec:noise_bound}

In practical LEO Doppler positioning, measurements are corrupted by noise disturbances. Under mild parameter perturbations, strong duality of SDP relaxations is preserved\cite{rosen_se-sync_2019}. Assuming the static ground receiver state remains unperturbed, measurement noise primarily affects the coefficient matrix $\mathbf{F}_\theta$ and gradient vector $\nabla f_\theta(\mathbf{y})$ in the objective function, leaving constraint equations $g_i(\mathbf{y}) = 0$ invariant.

To establish an analytical condition for relaxation tightness under noise, we apply the Abadie constraint qualification (ACQ) and Weyl's eigenvalue inequality\cite{cifuentes_local_2022}.

\begin{proposition}[Guaranteed Noise Bound]\label{prop:acq}
    Let $\overline{\mathbf{y}}$ denote the noiseless optimal state vector, $\sigma_N$ represent the $N$-th largest singular value of the constraint Jacobian matrix $\nabla \mathbf{g}(\overline{\mathbf{y}})$, and $\mathcal{G} = \sum_{i=1}^{2N} \mathbf{G}_i$. The SDP relaxation remains tight under noise perturbation if the parameter disturbance satisfies
    \begin{equation}
        \frac{1}{\sigma_N} \|\mathcal{G}\| \|\nabla f_\theta(\overline{\mathbf{y}})\| + \|\mathbf{F} - \mathbf{F}_\theta\| < \nu_{N+4}(\mathbf{F}_\theta)
        \label{equ:regular}
    \end{equation}
    where $\nu_{N+4}(\mathbf{F}_\theta)$ denotes the $(N+4)$-th smallest eigenvalue of the perturbed matrix $\mathbf{F}_\theta$.
\end{proposition}

Condition (\ref{equ:regular}) provides a theoretical bound on measurement noise levels for maintaining relaxation tightness. While optimality certification is evaluated post-optimization, inequality (\ref{equ:regular}) enables prior noise tolerance quantification in numerical simulations to evaluate system robustness.

\section{Experiments and Performance Evaluation}\label{sec:experiments}

This section evaluates the proposed initialization-free positioning framework through numerical simulations and real-world opportunistic LEO Doppler experiments.
All evaluated methods and hybrid pipelines are implemented in MATLAB R2024b using the SeDuMi solver via CVX\cite{boyd_convex_2004}:
\textbf{GN}, the standard Gauss-Newton solver using local linearization\cite{gratton_approximate_2007};
\textbf{DL}, Powell's Dog-Leg trust-region solver\cite{wright_numerical_1999};
\textbf{SDP}, the proposed convex relaxation solver providing an initialization-free cold-start estimate;
\textbf{{SDP-GN}}, the hybrid pipeline using the SDP estimate to seed the Gauss-Newton solver;
and \textbf{{SDP-DL}}, the hybrid pipeline using the SDP estimate to seed the Dog-Leg solver.

To improve solver numerical conditioning in the ECEF frame, receiver position coordinates are scaled by $10^{-7}$ and velocity/Doppler quantities by $10^{-3}$. Rank-tightness is confirmed when $\lambda_1(\mathbf{S}) / \lambda_2(\mathbf{S}) > 10^5$.
For solver configurations, the SDP solver via CVX is configured with high precision mode (\texttt{cvx\_precision high}) and an interior-point convergence tolerance of $\epsilon = 10^{-12}$ (\texttt{cvx\_solver\_settings('eps', 1e-12)}). For local search solvers (GN and DL), the maximum iteration count is set to $1000$ with a gradient and state update convergence tolerance of $10^{-6}$.

\begin{table}[htbp]
\centering
\caption{3D positioning error with initial points ranged from different initial distance in the simulation test.}
\label{tab:sim_initial}
\begin{tabular}{cccccc}
\hline
Init. Dist. [km]& GN& DL& SDP& SDP-GN&SDP-DL\\ \hline
1& 0.00& 0.00   & 0.71& 0.00&0.00\\
10& 0.00& 0.00   & 0.71 & 0.00&0.00\\
100& 0.00& 0.00   & 0.71 & 0.00&0.00\\
 580& 1839.60& 91.65&0.71& 0.00&0.00\\
1000& -& 892.23& 0.71 & 0.00&0.00\\ \hline
\end{tabular}
\end{table}

\begin{figure}
    \centering
  \includegraphics[width=1\linewidth]{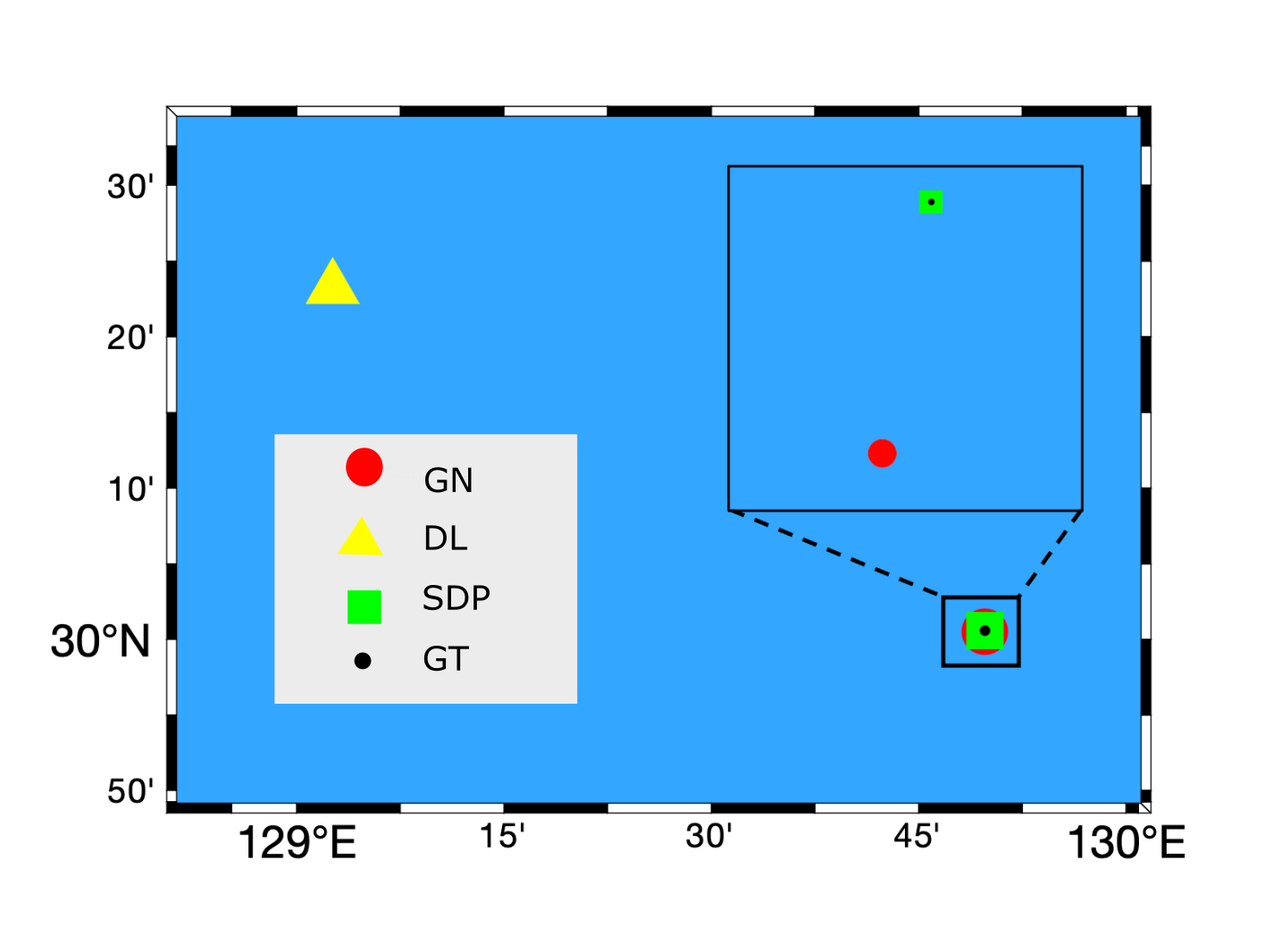}
    \caption{Estimated position with the initial distance of 580 km, where `GT' denotes ground-truth and other legends can be referred to Section \ref{sec:sim_settings}.}
    \label{fig:sim_initial}
\end{figure}

\subsection{Simulation Settings}\label{sec:sim_settings}

To evaluate the proposed cold-start positioning framework under ideal conditions, a constellation of 49 LEO satellites is synthesized at an altitude of $800 \text{ km}$, uniformly distributed over a $1500 \times 1500 \text{ km}$ spherical grid in the ECEF reference frame. Satellite orbital velocities are generated according to gravitational equations\cite{zhuang_gravity_2022}. The ground receiver is situated at the projection of the satellite network center on the Earth's surface, operating at a carrier frequency of $1.626 \text{ GHz}$. Monte Carlo simulations are repeated 40 times for each test scenario to evaluate both noiseless optimization behavior and parameter noise robustness.

\subsection{Performance in Noiseless Scenarios}\label{sec:sim_noiseless}

The positioning results under varying initial position offsets are summarized in Table \ref{tab:sim_initial}. When provided with accurate initial guesses (initial offset $\le 10 \text{ km}$), local search methods (GN and DL) achieve millimeter-level accuracy. However, as the initial position offset increases to $580 \text{ km}$, GN diverges with a positioning error of $1839.60 \text{ km}$, while DL yields an error of $91.65 \text{ km}$. At an initial offset of $1000 \text{ km}$, GN fails to output a feasible solution and DL error rises to $892.23 \text{ km}$. These results demonstrate that conventional local solvers depend strictly on precise initial estimates and frequently trap in local minima when initialized far from the true state.

In contrast, the proposed SDP method requires no prior initial position estimate and consistently converges to a bounded cold-start position with an average error of $0.71 \text{ km}$ across all initial offsets. This residual offset stems primarily from numerical precision limitations in interior-point SDP solvers under ECEF magnitude scaling. When seeded with the SDP output, the hybrid pipelines {SDP-GN} and {SDP-DL} successfully escape local minima traps, recovering exact positioning accuracy without requiring prior initial guesses, as illustrated in Fig. \ref{fig:sim_initial}.

\begin{figure}[htbp]
\centering
\begin{minipage}{0.48\linewidth}
\centering
\textbf{(a)}\\[2pt]
\includegraphics[width=\linewidth,height=0.22\textheight,keepaspectratio]{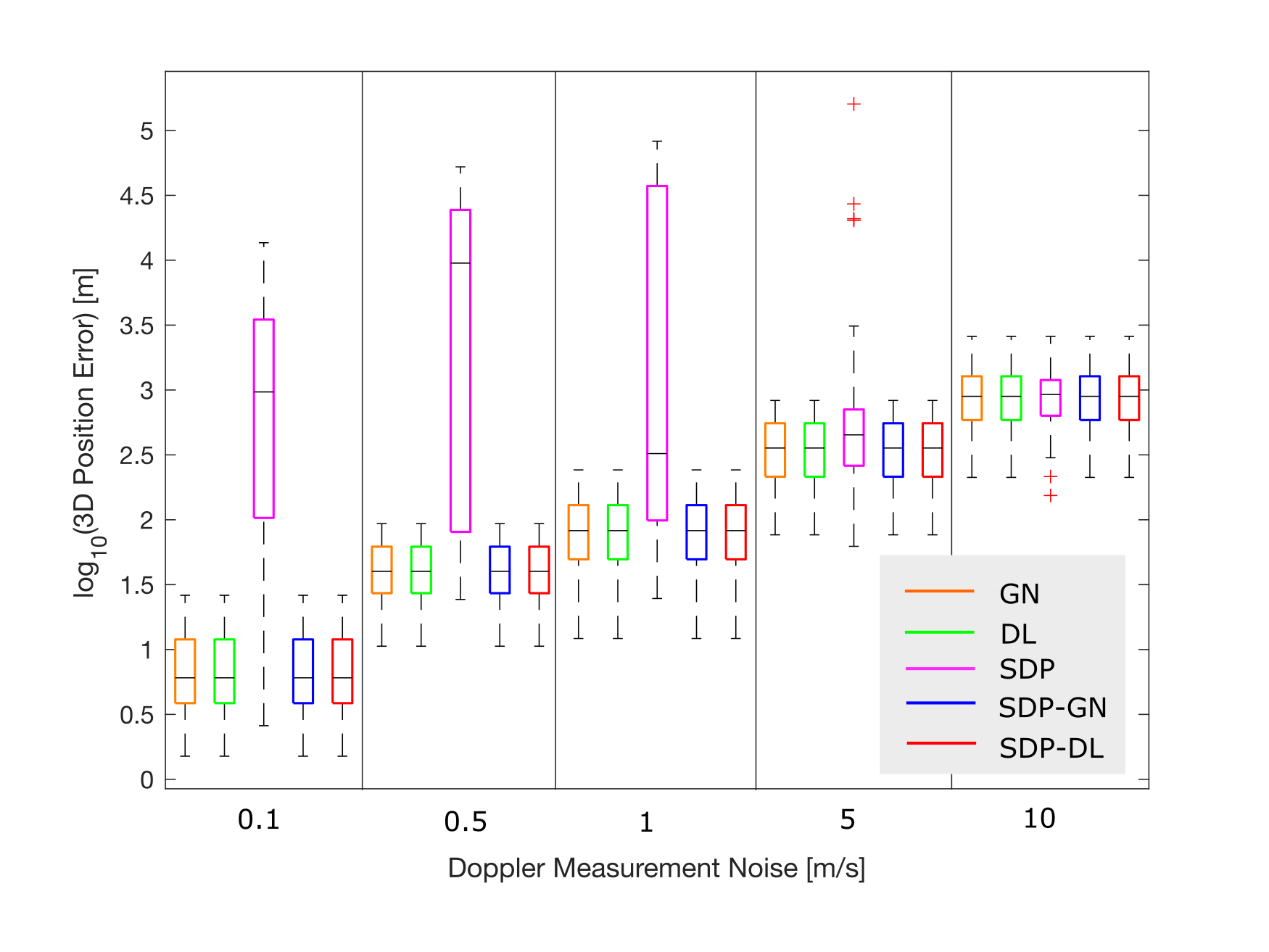}
\end{minipage}%
\hfill%
\begin{minipage}{0.48\linewidth}
\centering
\textbf{(b)}\\[2pt]
\includegraphics[width=\linewidth,height=0.22\textheight,keepaspectratio]{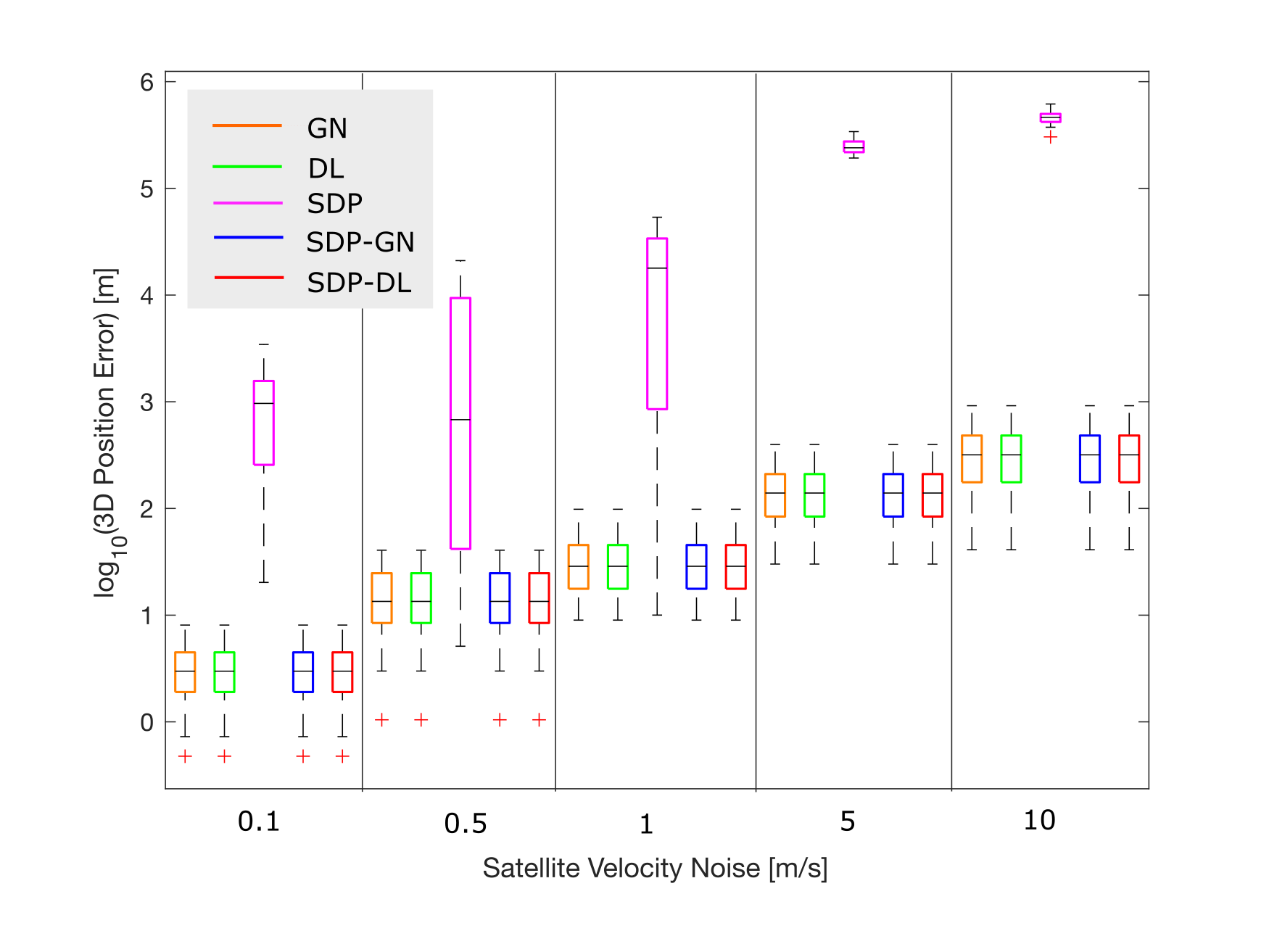}
\end{minipage}
\par\medskip
\begin{minipage}{0.65\linewidth}
\centering
\textbf{(c)}\\[2pt]
\includegraphics[width=\linewidth,height=0.22\textheight,keepaspectratio]{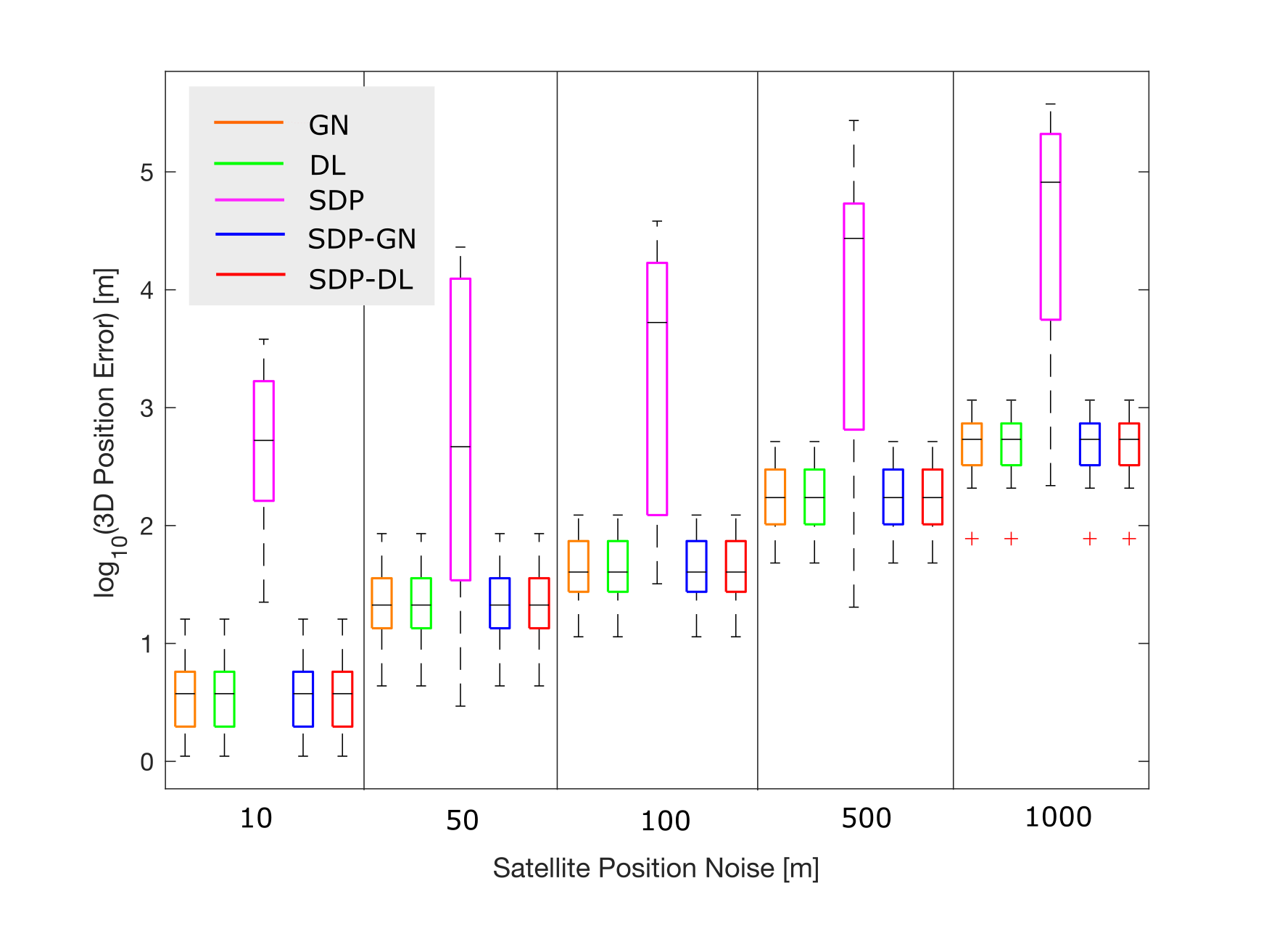}
\end{minipage}
\par
\caption{Positioning accuracy under parameter disturbances. (a) 3D position errors with various Doppler measurement noises. (b) 3D positioning errors with various satellite velocity noises. (c) 3D positioning errors with various satellite position noises.}
\label{fig:doppler_noise}
\label{fig:vel_noise}
\label{fig:pos_noise}
\end{figure}

% \subsection{Discussion: Performance and Limitation of the methods under noises}\label{sec:accuray}

\subsection{Discussion: Performance of Positioning in Noisy Cases}\label{sec:sim_noisy}

To test the robustness of our method, we add zero-mean Gaussian noise to Doppler shift data, satellite velocity vectors, and satellite positions. We test Doppler shift noise and satellite velocity noise at five standard deviation levels from $0.1 \text{ m/s}$ to $10 \text{ m/s}$. We also test satellite position noise with standard deviations from $10 \text{ m}$ to $1000 \text{ m}$.

The positioning errors are shown in Fig. \ref{fig:doppler_noise}(a--c). When using true positions as initial values, local methods GN and DL show similar accuracy at all noise levels. Although these noises grow the error in SDP, our SDP method still gives a stable and bounded initial position. When we use the SDP output to initialize local search methods, both {SDP-GN} and {SDP-DL} reach the same accuracy as local methods that use true initial values. This proves that our cold start method for LEO Doppler positioning handles parameter noise well in these noise disturbance tests.

\subsection{Prior Noise Bound for Rank-Tightness}\label{sec:bound}

\begin{figure}
    \centering
    \includegraphics[width=1\linewidth]{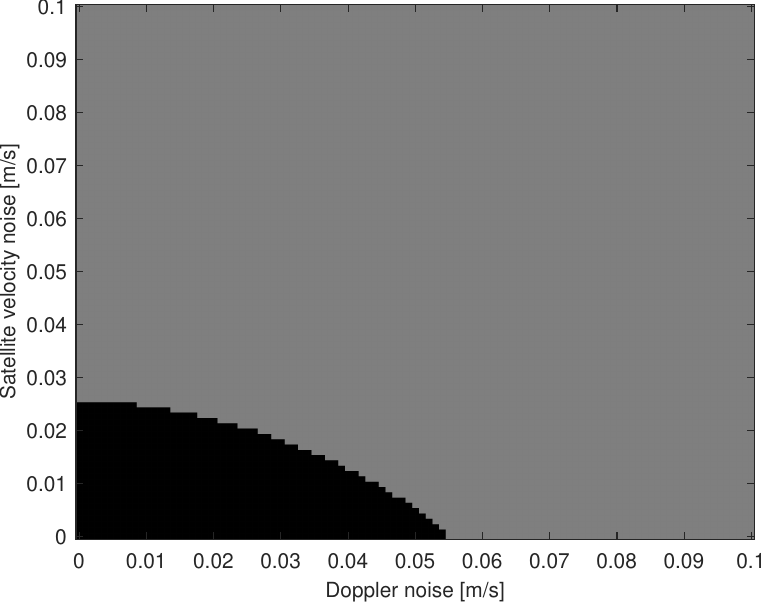}
    \caption{Demonstration of explicit noise bound, including satellite velocity and Doppler measurement noises, where grey cells denote rank-tightness and black cells denote explicit bounded noise.}
    \label{fig:noise_bound}
\end{figure}

To evaluate relaxation exactness prior to execution, we compute the theoretical sufficient noise bound for satellite velocity and Doppler shift measurements using the inequality (\ref{equ:regular}).
The predicted noise bounds are illustrated in Fig. \ref{fig:noise_bound}, where grey and black cells indicate rank-tightness, under which our method provide an optimal cold-start Doppler positioning solution. The black cells represent the theoretical noise limits predicted by Proposition \ref{prop:acq}, yielding a sufficient rank-tightness bound of approximately $2.5 \times 10^{-2} \text{ m/s}$ for satellite velocity noise and $5.5 \times 10^{-2} \text{ m/s}$ for Doppler measurement noise. Although conservative, this prior analysis enables users to quickly verify relaxation tightness before invoking time-intensive SDP optimization routines.

\begin{table}[htbp]
\caption{Key parameters in the real experiment.}
\begin{tabular}{ll}
\hline
Dataset             & LEO   Doppler positioning               \\ \hline
Constellation       & Iridium   (66 satellites in operation) \\
Orbit               & Polar   orbit $\sim$780 km altitude     \\
Frequency           & 1626270833   Hz                         \\
 Receiver Location&(22.3045966$\degree$, 114.180121$\degree$,61.384 m)\\
Visible   Satellite & 8                                       \\
Duration            & 35 minutes                                  \\
Measurements        & 436                                     \\
Ephemeris           & TLE+SGP4                                \\ \hline
\end{tabular}\label{tab:real_dataset}
\end{table}

\begin{figure}[htbp]
\centering
\begin{minipage}{0.48\linewidth}
\centering
\textbf{(a)}\\[2pt]
\includegraphics[width=\linewidth,height=0.22\textheight,keepaspectratio]{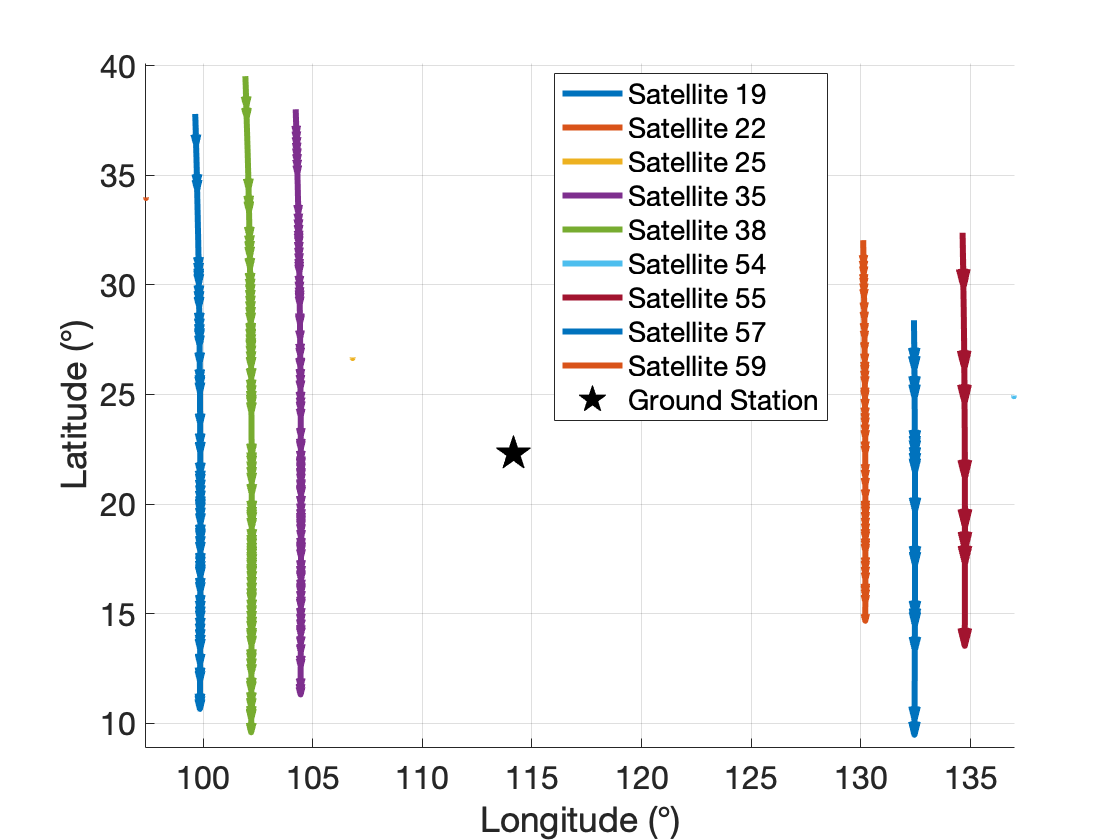}
\end{minipage}%
\hfill%
\begin{minipage}{0.48\linewidth}
\centering
\textbf{(b)}\\[2pt]
\includegraphics[width=\linewidth,height=0.22\textheight,keepaspectratio]{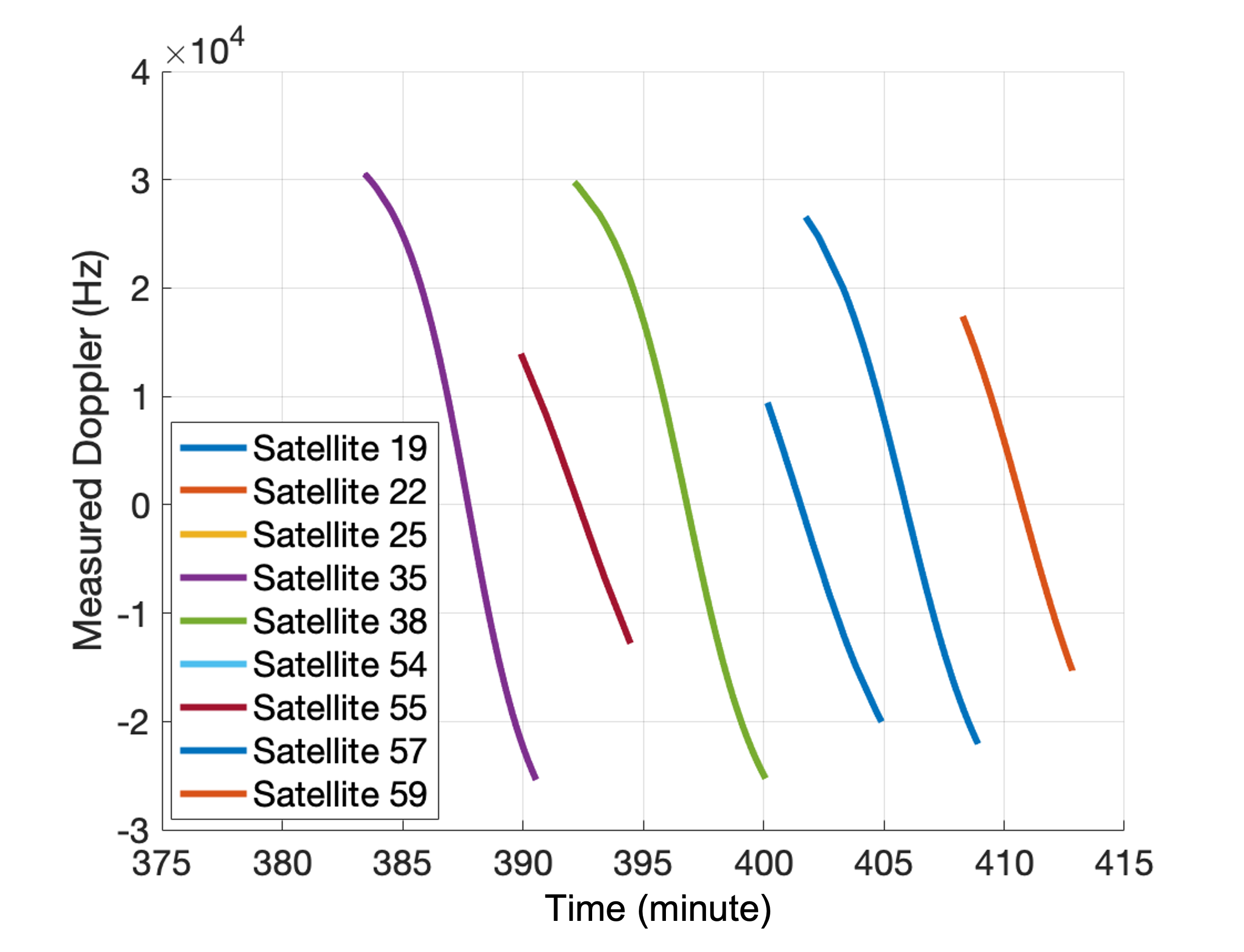}
\end{minipage}
\par\medskip
\begin{minipage}{0.72\linewidth}
\centering
\textbf{(c)}\\[2pt]
\includegraphics[width=\linewidth,height=0.25\textheight,keepaspectratio]{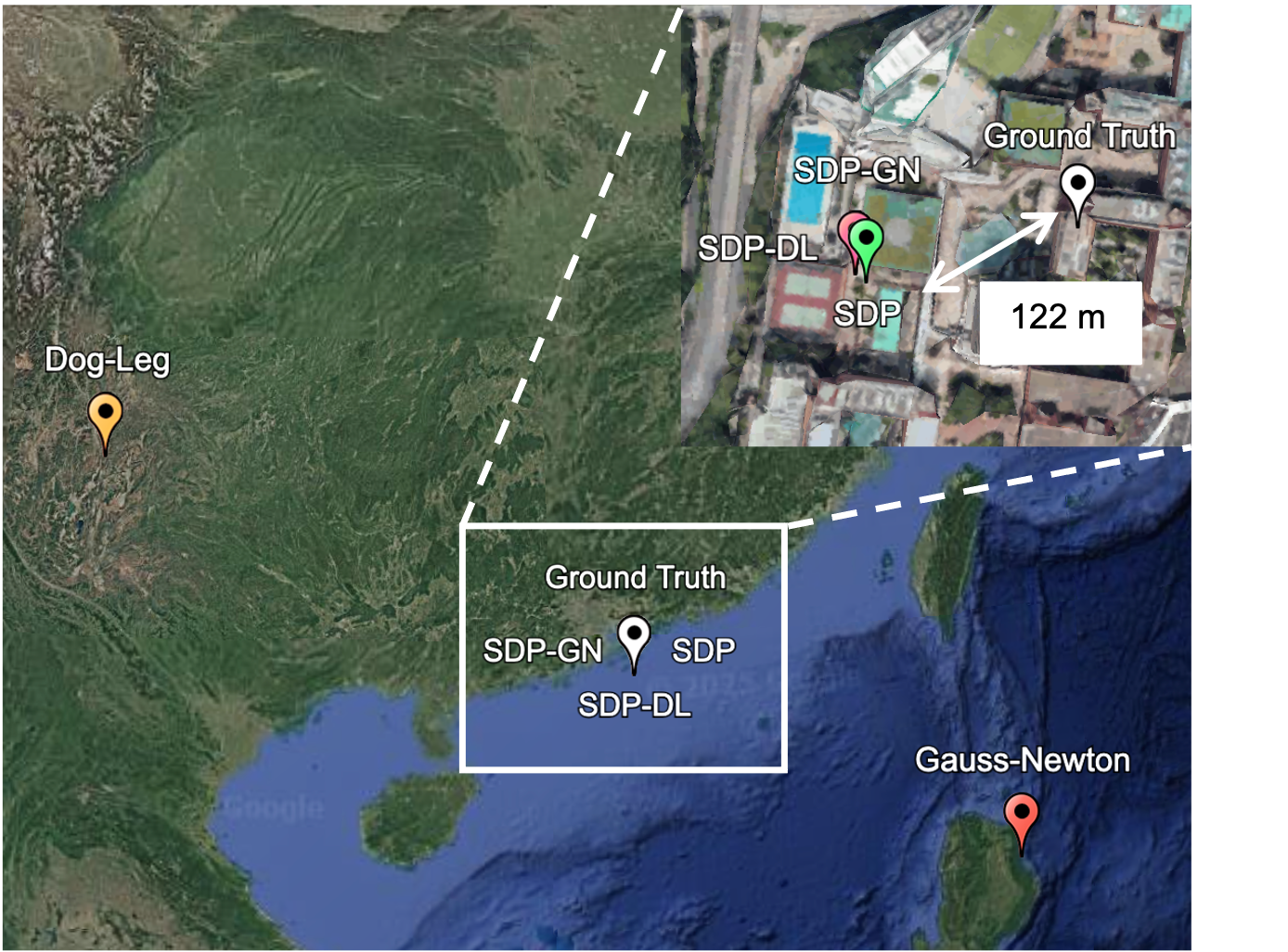}
\end{minipage}
\par
\caption{Real-world Iridium-NEXT experiment. (a) Illustration of satellite orbit determination using Two-Line Element (TLE) data and the Simplified General Perturbations Satellite Orbit Model 4 (SGP4) propagation model. The SGP4 model calculates time-varying satellite positions and velocities from TLE sets to provide reference orbital parameters for LEO Doppler positioning. (b) Doppler shift measurements in the real test. (c) Comparison between local minima and global minimum on Google Earth under a 1000 km initial offset. Local search methods get trapped in local optima due to large initial errors, whereas the proposed SDP solver reliably calculates a cold-start initial position with a 3D error of 140 m and a 2D horizontal error of 122 m.}
\label{fig:real_eph}
\label{fig:real_dopp}
\label{fig:real_google}
\end{figure}

\subsection{Real-World Opportunistic LEO Doppler Positioning}\label{sec:real_test}

To validate practical operational efficacy, the proposed method is evaluated using real-world opportunistic Iridium Doppler shift observations collected via a software-defined radio (SDR) setup on the campus of The Hong Kong Polytechnic University. Key dataset parameters are detailed in Table \ref{tab:real_dataset}. Over a tracking interval of 35 minutes, 436 Doppler shift measurements were extracted from 8 visible satellites. Satellite position and velocity trajectories derived from Two-Line Element (TLE) ephemerides using the Simplified General Perturbations Satellite Orbit Model 4 (SGP4) model are illustrated in Fig. \ref{fig:real_eph}, and the corresponding Doppler shift measurements are plotted in Fig. \ref{fig:real_dopp}.

The positioning performance under varying initial position offsets is presented in Table \ref{tab:real_initial}. When provided with a precise initial guess (initial distance $\le 100 \text{ km}$), local search methods (GN and DL) achieve a 3D positioning error of $0.13 \text{ km}$. However, when the initial position offset increases to $1000 \text{ km}$ or larger, both local solvers fail to converge to the true position. Specifically, GN diverges or fails to yield a feasible output, while DL exhibits a severe 3D error up to $8445.10 \text{ km}$, confirming that local search routines are prone to local minima in cold-start scenarios.

In contrast, the proposed SDP method requires no prior initial position knowledge and consistently solves the relaxed problem, outputting a stable cold-start position with a 3D error of $0.14 \text{ km}$. When seeded with the SDP estimate, the hybrid pipelines {SDP-GN} and {SDP-DL} successfully eliminate local search divergence, matching the optimal accuracy of baseline local solvers with a 3D error of $0.13 \text{ km}$ (corresponding to a horizontal error of approximately $122 \text{ m}$ from the ground truth). These real-world results align with simulation findings, confirming that the proposed SDP framework serves as an effective, initialization-free cold-start engine for opportunistic LEO Doppler navigation. In terms of computational overhead, solving the SDP relaxation on 436 Iridium-NEXT observations takes approximately $39.69 \text{ s}$. While this processing time reflects the complexity of interior-point optimization on high-dimensional lifted matrices, it represents a one-time execution cost solely incurred during complete cold start. Once seeded with the SDP solution, the local refinement stage via {SDP-GN} or {SDP-DL} completes in less than $0.008 \text{ s}$ ($8 \text{ ms}$), providing an efficient pipeline that combines guaranteed global initialization with real-time local tracking.

\begin{table}[htbp]
    \centering
    
    \caption{3D positioning errors of the compared methods using different initial distances in the real test.}
\begin{tabular}{cccccc}
\hline
Init dist {[}km{]} & GN   & DL     & SDP  & SDP-GN & SDP-DL \\ \hline
10                       & 0.13 & 0.13   & 0.14 & 0.13   & 0.13   \\
100                       & 0.13 & 0.13   & 0.14 & 0.13   & 0.13   \\
500                       & 0.13 & 0.13   & 0.14 & 0.13   & 0.13   \\
1000                       & 2342.50  & 2069.00 & 0.14 & 0.13   & 0.13   \\
5000                       & -  & 8445.10 & 0.14 & 0.13   & 0.13   \\ \hline
\end{tabular}\label{tab:real_initial}
\end{table}

\section{Conclusion}\label{sec:conclusion}

In this paper, we proposed an initialization-free LEO Doppler positioning framework using SDP relaxation to address the cold-start challenge in GNSS-denied IoT environments. By reformulating the non-convex formulation into a convex SDP, the proposed method provides a reliable receiver position estimate without requiring prior location knowledge. We also derived theoretical duality tightness conditions and noise bounds under parameter perturbations. Simulations and real-world experiments using Iridium-NEXT LEO signals confirm that the SDP output serves as a high-quality cold-start initializer for local solvers, refining positioning accuracy down to 130 m. All code and datasets are made open source. Future work will extend the framework to dynamic navigation and joint orbital parameter estimation via sum-of-squares hierarchies.

\section*{Data availability}
The data supporting the findings of this study are publicly available at \url{https://github.com/Baoshan-Song/Certifiable-Doppler-positioning}, which provides synthetic and real-world examples, including Iridium Doppler positioning data.

\section*{Code availability}
The implementation of the semidefinite-relaxation positioning method and example scripts are available at \url{https://github.com/Baoshan-Song/Certifiable-Doppler-positioning}. The implementation uses MATLAB and the CVX convex optimisation package.

\backmatter

\bmhead{Author contributions}
B. S. designed the semidefinite programming (SDP) method. W. W. and L. H. formulated the LEO positioning model. B. X. and Q. Z. provided the open-source data. P. Y. and Y. L. reviewed the manuscript.

\bibliography{sn-sr}

\end{document}